\documentclass[11pt]{article}

\usepackage[preprint]{acl}

\usepackage{times}
\usepackage{latexsym}

\usepackage[T1]{fontenc}

\usepackage[utf8]{inputenc}

\usepackage{microtype}

\usepackage{inconsolata}

\usepackage{graphicx}

\usepackage{amsmath}
\usepackage{times}
\usepackage{framed}
\usepackage{ragged2e}
\usepackage{bm}
\usepackage{soul}
\usepackage{latexsym}
\usepackage{subfigure}
\usepackage{amsthm}
\usepackage{graphicx}
\usepackage{float}
\usepackage{longtable}
\usepackage{booktabs} 
\usepackage{multirow}
\usepackage{multicol}
\usepackage{amssymb}
\usepackage{dashbox}%
\usepackage{multirow}
\usepackage{textcomp}
\usepackage{tcolorbox}
\usepackage{bbding}
\usepackage{bbm}
\usepackage[ruled, vlined, linesnumbered]{algorithm2e}
\usepackage{mathtools}
\usepackage{adjustbox}
\usepackage[ruled,vlined]{algorithm2e}
\usepackage{color, soul}
\usepackage{pifont}
\usepackage{array}
\newcolumntype{P}[1]{>{\centering\arraybackslash}p{#1}}
\newcolumntype{M}[1]{>{\centering\arraybackslash}m{#1}}
\usepackage{cleveref}
\crefname{section}{§}{§§}
\Crefname{section}{§}{§§}
\crefname{figure}{Figure}{Figure}
\Crefname{figure}{Figure}{Figure}
\crefname{table}{Table}{Table}
\Crefname{table}{Table}{Table}
\usepackage{tabularx}
\usepackage{booktabs}
\usepackage[table,xcdraw]{xcolor}
\usepackage{makecell}

\newcommand\ourmodel{\textsc{WildReward}\xspace}
\newcommand\ourmodelsmall{\textsc{WildReward-4B}\xspace}
\newcommand\ourmodellarge{\textsc{WildReward-8B}\xspace}
\newcommand\ourdata{\textsc{WildFB}\xspace}

\usepackage{tcolorbox}
\tcbuselibrary{listings, breakable} 

\title{\ourmodel: Learning Reward Models from \\ In-the-Wild Human Interactions}

\author{Hao Peng\thanks{\quad Work done during an internship at Zhipu AI}, Yunjia Qi, Xiaozhi Wang, Zijun Yao, Lei Hou, Juanzi Li\thanks{\quad Corresponding author: Juanzi Li}\\
Department of Computer Science and Technology, Tsinghua University \\
\texttt{\{peng-h24\}@mails.tsinghua.edu.cn}}

\begin{document}
\maketitle
\begin{abstract}

Reward models (RMs) are crucial for the training of large language models (LLMs), yet they typically rely on large-scale human-annotated preference pairs.
With the widespread deployment of LLMs, in-the-wild interactions have emerged as a rich source of implicit reward signals. 
This raises the question: \textit{Can we develop reward models directly from in-the-wild interactions?} 
In this work, we explore this possibility by adopting WildChat as an interaction source and proposing a pipeline to extract reliable human feedback, yielding 186k high-quality instances for training \ourmodel via ordinal regression directly on user feedback without preference pairs.
Extensive experiments demonstrate that \ourmodel achieves comparable or even superior performance compared to conventional reward models, with improved calibration and cross-sample consistency. We also observe that \ourmodel benefits directly from user diversity, where more users yield stronger reward models. Finally, we apply \ourmodel to online DPO training and observe significant improvements across various tasks.
Code and data are released at \url{https://github.com/THU-KEG/WildReward}.



\end{abstract}

\section{Introduction}

Reward models (RMs) are crucial for the training and inference-time scaling of large language models (LLMs). They are typically used to model human preferences and trained on large-scale human-annotated preference pairs~\citep{ouyang2022training}. Prior work has primarily focused on collecting preference pairs~\citep{wang2024helpsteer, wang2025helpsteer3,liu2025skywork}, requiring substantial annotation efforts.

\begin{figure}[t]
    \centering

    \includegraphics[width=0.99\linewidth]{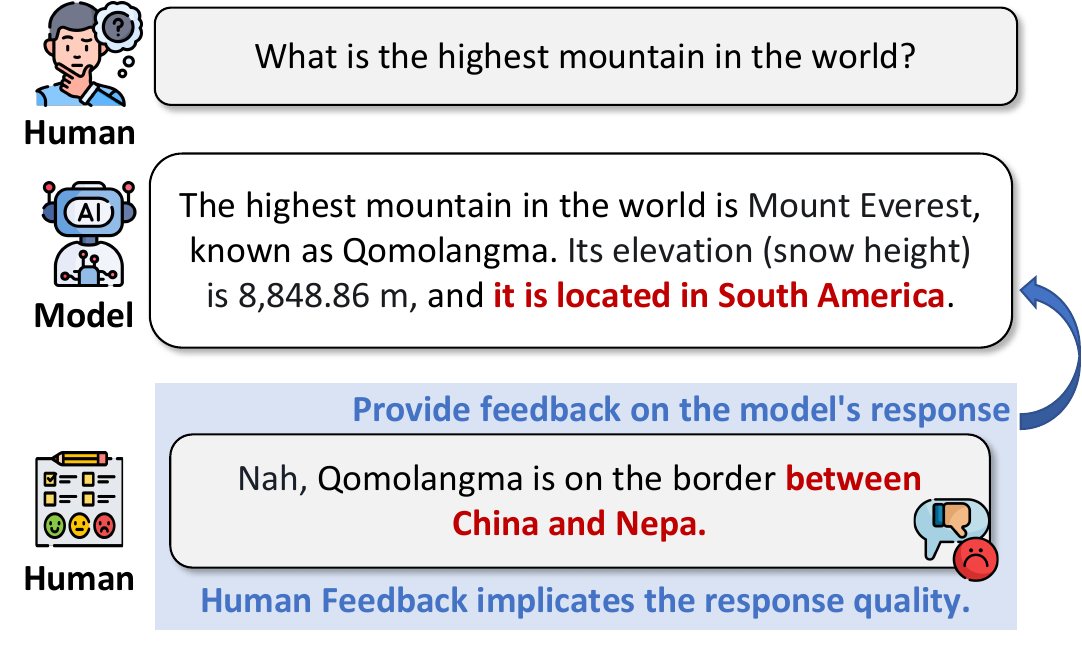} 
    \caption{
     Illustration of a human-LLM interaction with implicit feedback signals in the conversation. The user provides valid feedback and identifies an error.
    }
    \label{fig:fig1}
\end{figure}

With the widespread deployment of LLMs, numerous in-the-wild interactions with humans have emerged, such as human-LLM conversations~\citep{zhaowildchat24,zhenglmsys24}. These interactions typically contain rich human feedback. For example, as shown in Figure~\ref{fig:fig1}, humans provide textual feedback regarding the previous model response. This feedback directly reflects the response quality and authentic human preferences, thereby naturally serving as training data for reward models. Despite the large scale and richness of these interactions, their utility remains under-explored. This raises a critical question: \textit{Can we develop reward models directly from these in-the-wild interactions?} 

In this work, we explore training reward models using in-the-wild human interactions. Specifically, we leverage WildChat~\citep{zhaowildchat24}, a large-scale human-LLM conversation dataset. We first conduct a preliminary analysis of user queries within WildChat, which reveals two primary observations and challenges: (1) \textbf{Feedback sparsity}. Feedback is mostly implicit. Approximately $82\%$ of follow-up queries do not explicitly convey feedback or preference regarding the previous response. Notably, explicit positive feedback is particularly scarce, which accounts for only $1\%$. (2) \textbf{Feedback noise}. User feedback is prone to noise, especially in safety scenarios. For instance, when an LLM correctly refuses a sensitive question, the user may provide negative feedback, which is unjustified. To mitigate these issues, we propose an automated pipeline to filter noise and extract reliable feedback. Specifically, we classify user feedback into five levels of satisfaction, indicating response quality, including explicit rejection, error correction, neutral ambiguity, positive engagement, and explicit satisfaction. We adopt gpt-oss-120b~\citep{agarwal2025gpt} for automatic classification and adopt a conservative strategy that defaults to neutral ambiguity in the absence of strong evidence to minimize label noise. Furthermore, we propose a two-stage refinement strategy to extract implicit feedback and mitigate noise: (1) Implicit feedback mining, which recovers implicit positive signals; (2) Refusal validation, which validates justified safety refusals. Finally, we exclude the neutral ambiguity subset, resulting in \ourdata, which contains 186k instances, each consisting of a conversation history, a user query, a response, and a label indicating the response quality. We verify data quality through sampled 100 instances and observe little noise. For training reward models, we adopt the ordinal regression objective~\citep{wang2025survey} to explicitly learn accurate relative rankings of user feedback, resulting in our reward model \ourmodel.

We conduct extensive experiments to validate the efficacy of \ourmodel. We first evaluate \ourmodel on standard and widely used reward model benchmarks, including RewardBench~\citep{lambert2025rewardbench}, RM-Bench~\citep{liurm25}, PPE~\citep{frickevaluate25}, and JudgeBench~\citep{tanjudgebench25}. We find that \ourmodel achieves performance comparable or superior to conventional reward models. This demonstrates that \ourmodel effectively captures general human preferences without any dedicated human-annotated preference pairs.
We conduct extensive analyses and draw the following conclusions:
(1) Data strategy. Both implicit feedback mining and refusal validation strategies are beneficial. Furthermore, \ourmodel benefits from user diversity, as more users yield stronger models.
(2) Calibration. \ourmodel is well-calibrated. By using the score margin between chosen and rejected responses as a proxy for confidence, we observe a strong positive correlation between confidence and accuracy. This indicates that \ourmodel can be integrated with more powerful LLMs~\citep{xuask25} or humans to produce more accurate reward signals.
(3) Cross-sample consistency. \ourmodel exhibits strong global score calibration and provides a unified and meaningful score for assessing response quality.
We introduce an implicit feedback prediction task designed to predict binary user reception (positive or negative) based on the conversation context, user query, and response. \ourmodel achieves significantly higher ROC-AUC scores.
(4) Application in DPO training. \ourmodel effectively guides policy model training. When applied to online DPO~\citep{rafailov2023direct}, the trained model achieves significant improvements across various downstream tasks, including mathematical reasoning, instruction following, and creative writing.
In conclusion, this work demonstrates the potential of real-world interactions and highlights a promising direction for leveraging these rapidly growing resources. We encourage more research efforts to explore this area in the future.

\section{Methodology}
We adopt \textbf{WildChat}~\citep{zhaowildchat24}, a human-LLM conversation dataset, as primary interaction source. This section presents a preliminary analysis of WildChat (\cref{sec:pre_analysis}), the automated pipeline used to extract feedback and construct a high-quality dataset \textbf{\ourdata} (\cref{sec:data_construct}), and the training method (\cref{sec:training}) for the reward model \textbf{\ourmodel}.

\begin{figure*}[t]
    \centering
    \includegraphics[width=1.0\linewidth]{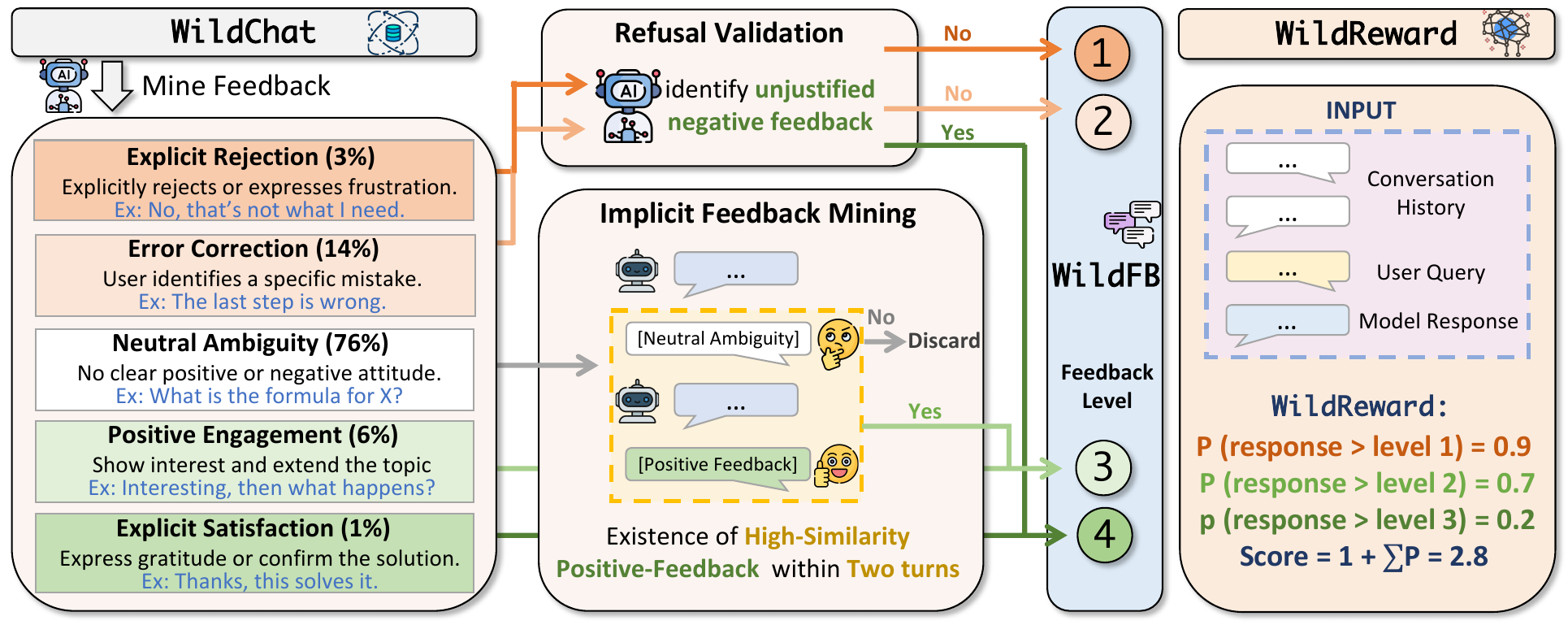} 
    \caption{Overview of the proposed pipeline for extracting human feedback from in-the-wild conversations.
    }
    \label{fig:pipeline}
\end{figure*}

\subsection{Preliminary Analysis}
\label{sec:pre_analysis}

We conduct a preliminary analysis of WildChat. Specifically, we first sample $10,000$ instances, each consisting of a conversation history, the user query, the corresponding model response, and the user's follow-up query. 
We then analyze the follow-up query, as it potentially reflects the user preference or the quality of the preceding response. 
We adopt gpt-oss-120b~\citep{agarwal2025gpt} to automatically classify the follow-up query into three feedback categories: \textit{Negative}, \textit{Neutral}, and \textit{Positive}. We observe that valid feedback is sparse: approximately $82\%$ of queries are \textit{Neutral}, i.e., do not explicitly express feedback, while $17\%$ contain negative feedback, e.g., pointing out errors. Only $1\%$ are \textit{Positive}, which is expected as users rarely express explicit gratitude in natural interactions.

We further sample $200$ instances from this classified set and conduct a manual inspection. In the \textit{Neutral} category, approximately $86\%$ are new requests unrelated to the previous response, while the remaining $14\%$ are relevant follow-up questions, which may serve as implicit feedback. 
In the \textit{Negative} and \textit{Positive} categories, the classification accuracy of gpt-oss-120b is high, where most follow-up queries express corresponding negative or positive feedback. However, we still find a type of noise: an LLM correctly refuses a sensitive question, yet the user provides negative feedback, which is unjustified.
In conclusion, while in-the-wild conversations contain valuable feedback signals, they present challenges regarding sparsity and noise. It requires a well-designed pipeline for extracting valid feedback. More details are in Appendix~\ref{sec:app_data_construct}.

\subsection{\ourdata Dataset Construction}
\label{sec:data_construct}

Based on the preliminary analysis, we propose an automated pipeline to extract valid feedback and mitigate noise from WildChat. The overall framework is shown in Figure~\ref{fig:pipeline}. To capture fine-grained feedback signals, we classify user feedback into five levels of satisfaction, indicating response quality, and the corresponding descriptions are shown in Figure~\ref{fig:pipeline}. Inspired by our initial findings in \cref{sec:pre_analysis} that relevant follow-up questions may signal active user involvement, we consider such engagement as a form of implicit positive feedback. Therefore, we introduce the \textit{Positive Engagement} category to collect more instances with positive feedback. 
Given the reliability of gpt-oss-120b verified in \cref{sec:pre_analysis}, we employ it for automatic feedback classification. To minimize label noise, we adopt a conservative strategy that defaults to \textit{Neutral Ambiguity} in the absence of strong evidence. The distribution of extracted feedback is shown in Figure~\ref{fig:pipeline}.

To further extract valid feedback and mitigate labeling noise, we propose a two-stage refinement strategy:
(1) \textbf{Implicit feedback mining}. As observed in \cref{sec:pre_analysis}, the \textit{Neutral Ambiguity} category contains implicit positive feedback. To exploit this, we find that if a user provides positive feedback in adjacent turns within a coherent conversation context, e.g., the same topic, the intermediate responses are likely also 
of high quality with positive feedback. We analyze 20 randomly sampled instances and find that $90\%$ support this intuition. 
Consequently, we mine \textit{Neutral Ambiguity} instances where the user query shares high semantic similarity ($>0.6$) with a positive-feedback query within a two-turn window. The semantic similarity is computed using the cosine similarity of sentence embeddings derived from all-MiniLM-L6-v2\footnote{\url{https://huggingface.co/sentence-transformers/all-MiniLM-L6-v2}}. These instances are reclassified as positive feedback, yielding approximately $12,310$ additional samples and expanding the positive feedback subset by $29\%$.
(2) \textbf{Refusal Validation}. As observed in \cref{sec:pre_analysis}, negative user feedback in safety contexts may introduce noise, as users often respond negatively even when the model correctly refuses sensitive queries. To address this, we employ gpt-oss-120b to analyze instances of the \textit{Explicit Rejection} and \textit{Error Correction} categories, determining whether the negative feedback is unjustified given the valid refusal, which finally fixes about $572$ such errors. Although the scale is small, this correction significantly improves performance on safety subsets in reward model benchmarks, as demonstrated in our ablation study (\cref{sec:ablation}).
Finally, we \textbf{exclude} the remaining \textit{Neutral Ambiguity} subset as it lacks clear feedback signals and obtain \textbf{\ourdata}, a high-quality dataset comprising approximately 186k instances across four feedback categories, each containing a conversation history, a user query, a response, and a label indicating the response quality. More data construction details are placed in Appendix~\ref{sec:app_data_construct}.


\subsection{Training \ourmodel}
\label{sec:training}

Conventional reward models typically rely on a large-scale set of preference pairs and are trained using the Bradley-Terry model~\citep{bradley1952rank}. In contrast, \ourdata consists of point-wise user feedback labels that exhibit an intrinsic ordinal ranking structure. 
For instance, \textit{Positive Engagement} reflects a higher level of user satisfaction or response quality than the \textit{Error Correction} category. 
To leverage this ranking structure, we map the four feedback categories (excluding the \textit{Neutral Ambiguity} subset) to discrete quality scores ranging from \textit{1} to \textit{4}, where \textit{1} represents \textit{Explicit Rejection} and \textit{4} represents \textit{Explicit Satisfaction}. We train a reward model to learn this hierarchy, enabling it to produce reliable ranking scores for model responses. Specifically, we adopt an ordinal regression objective for training, which is demonstrated effective to explicitly learn the accurate relative rankings~\citep{wang2025survey}. Compared to standard regression, ordinal regression models the inherent order of feedback without assuming uniform intervals and provide probabilistic outputs for confidence filtering.

Formally, given a training instance consisting of a dialogue history $c$, a user query $q$, the corresponding response $s$, and an associated feedback label $y \in \{1,2,3,4\}$, we define the model input as $x=(c,q,s)$. The reward model is trained by minimizing the following objective:

\begin{small}
\begin{align*}
\mathcal{L} = - \sum_{k=1}^{K-1} \big[
    &\mathbb{I}(y>k) \log P(y>k | x; \theta) \\
    & + (1-\mathbb{I}(y>k)) \log(1-P(y>k | x; \theta))
\big]
\end{align*}
\end{small}

Here, $\theta$ represents the learnable parameters of the reward model. 
$K$ is the total number of ordinal categories, which is $4$ in this case. $\mathbb{I}$ denotes the indicator function. During inference, we compute a continuous reward score for each input $x$. The final reward score is computed as follows:

\begin{small}
\begin{equation*}
R(x) = 1 + \sum_{k=1}^{K-1} P(y > k | x; \theta)
\end{equation*}
\end{small}

This score represents the expected value of the predicted feedback. As a probabilistic measure, it can be effectively used for confidence filtering.

\section{Experiment}

\begin{table*}[t]
    \centering
    \small
    \resizebox{\linewidth}{!}{
    \begin{tabular}{lrrrrrrrr}\\
    \toprule
    \multirow{2}{*}{Model} & \multirow{2}{*}{RewardBench} & \multicolumn{3}{c}{RM-Bench} & \multicolumn{2}{c}{PPE} & \multirow{2}{*}{JudgeBench} \\
    \cmidrule(lr){3-5} \cmidrule(lr){6-7}
    & & Easy & Normal & Hard & Human & Correctness & \\
    \midrule
    ArmoRM-Llama3-8B-v0.1 & $90.4$ & $80.4$ & $71.5$ & $55.8$ & $60.6$ & $60.6$ & $59.7$ \\
    Athene-RM-8B & $84.8$ & $89.8$ & $76.6$ & $51.4$ & $\mathbf{64.6}$ & $62.0$ & $70.1$ \\
    Llama-3-OffsetBias-RM-8B & $89.0$ & $83.9$ & $73.2$ & $56.9$ & $59.2$ & $64.1$ & $63.5$ \\
    Skywork-Reward-Llama-3.1-8B-v0.2 & $93.1$ & $70.5$ & $74.2$ & $49.3$ & $62.2$ & $62.5$ & $62.9$  \\
    Internlm2-20b-reward & $90.2$ & $79.4$ & $74.2$ & $62.8$ & $61.0$ & $63.0$ & $ 64.3$ \\
    Skywork-Reward-Gemma-2-27B-v0.2 & $94.3$ & $88.9$ & $71.9$ & $42.1$ & $63.6$ & $61.9$ & $66.5$ \\
    Llama-3.1-Nemotron-70B & $93.9$ & $\mathbf{92.2}$ & $76.5$ & $47.8$ & $64.2$ & $63.2$ & $65.8$ \\
    INF-ORM-Llama3.1-70B & $\mathbf{95.1}$ & $92.1$ & $\mathbf{80.0}$ & $54.0$ & $64.2$ & $64.4$ & $\mathbf{70.2}$ \\
    \midrule
    \ourmodelsmall & $83.6$ & $82.0$ & $77.0$ & $68.6$ & $61.6$ & $63.6$ & $61.1$ \\
    \ourmodellarge & $86.0$ & $83.5$ & $78.4$ & $\mathbf{69.7}$ & $62.5$ & $\mathbf{65.6}$ & $66.0$ \\
    \bottomrule
    \end{tabular}
    }
    \caption{Experimental results (\%) on several representative reward model benchmarks. PPE Human and Correctness denote the human and correctness preference subset, respectively. The highest score in each column is in \textbf{bold}.}
    \label{tab:rewardbench_results}
\end{table*}

This section presents the experimental setup (\cref{sec:exp_setup}) and evaluation results on standard reward benchmarks (\cref{sec:rm_results}). We further analyze our data construction strategy (\cref{sec:ablation}), investigate the properties of \ourmodel (\Cref{sec:calibration,sec:consistency}), and demonstrate its effectiveness in guiding DPO training (\cref{sec:dpo_training}).



\subsection{Experimental Setup}
\label{sec:exp_setup}

Regarding \textbf{implementation details}, we use Qwen3 4B and 8B~\citep{yang2025qwen3} as backbone LLMs. We train these models on \ourdata for one epoch to develop the reward models \ourmodelsmall and \ourmodellarge, with a training batch size of $512$ and a learning rate of $1 \times 10^{-5}$.
Regarding investigated \textbf{baselines}, we adopt various representative reward models for comparison, which are typically trained on large-scale preference pairs, including Llama-3-OffsetBias-RM-8B~\citep{park2024offsetbias}, ArmoRM~\citep{wang2024interpretable}, Athene-RM~\citep{Athene2024}, Skywork-Reward~\citep{liu2024skywork}, InternLM2-Reward~\citep{cai2024internlm2}, INF-ORM-Llama3.1-70B~\citep{INF-ORM-Llama3.1-70B}, and Llama-3.1-Nemotron-70B~\citep{wang2024helpsteer2preferencecomplementingratingspreferences}.
Regarding \textbf{evaluation benchmarks}, we adopt standard and widely used reward model benchmarks, including RewardBench~\citep{lambert2025rewardbench}, RM-Bench~\citep{liurm25}, PPE~\citep{frickevaluate25}, and JudgeBench~\citep{tanjudgebench25}. 
We leverage all three difficulty levels (easy, normal, hard) of RM-Bench. RewardBench, RM-Bench, PPE Human, and JudgeBench use a binary choice setting to select the chosen response, while PPE Correctness employs a Best-of-N evaluation setting, which aligns with test-time scaling evaluation. The evaluation benchmarks cover diverse domains, including creative writing, instruction following, mathematics, commonsense reasoning, coding, and safety.

\subsection{Reward Model Benchmarking Result}
\label{sec:rm_results}

The experimental results are presented in Table~\ref{tab:rewardbench_results}, we can observe that:
(1) \ourmodel achieves comparable or even superior performance to conventional reward models without human-annotated preference pairs. Notably, \ourmodel, with only 4 or 8 billion parameters, surpasses the performance of much larger 70B reward models. It demonstrates that we can train reward models directly from human feedback, rather than relying on preference pairs and also confirms that there are valid human feedback exists within in-the-wild interactions.
Furthermore, this approach is inherently data scalable, given the vast amount of such interaction data available in the real world. One may collect more human feedback and develop more advanced reward models. (2) On RM-Bench Hard and PPE Correctness, \ourmodel demonstrates superior performance. RM-Bench Hard specifically evaluates robustness to superficial cues, such as irrelevant styles and length bias, and the ability to select factual responses~\citep{liurm25}. PPE Correctness also evaluates objective factual accuracy. The superior results on these two benchmarks demonstrate that \ourmodel is more robust to superficial biases. This observation is reasonable, as humans typically provide negative feedback to verbose yet incorrect answers in real-world interactions.
(3) \ourmodellarge consistently outperforms \ourmodelsmall, which indicates that larger models can more effectively leverage in-the-wild interaction data. This trend also demonstrates the effectiveness of \ourdata and suggests the potential for model scaling. However, due to computational constraints, we leave the exploration of further scaling to future work.
In conclusion, the experimental results demonstrate the significant potential of training reward models from in-the-wild human interactions. It also validates that vast amounts of human interactions are a potentially vital resource for the future of LLM training.


\subsection{Analysis on Data Strategy}
\label{sec:ablation}

We analyze the impact of different data curation strategies on the resulting reward models. We first conduct an ablation study to evaluate our proposed data strategies described in \cref{sec:data_construct}: implicit feedback mining and refusal validation. Specifically, we exclude the corresponding data and retrain the model, while keeping all other configurations identical.
The results are presented in Table~\ref{tab:ablation}.
``SRF'' denotes the Safety-Refusal subset, where the prompt is sensitive and the refusal response is the chosen one. ``SRP'' denotes the Safety-Response subset, where the prompt is normal and the standard, helpful response is the chosen one. 
We can observe that ablating either data strategy results in a significant performance degradation, which demonstrates that both the two strategies are effective for collecting high-quality data. 
The impact is particularly significant on the safety subset, where SRF performance drops by $60\%$ when the Refusal Validation strategy is excluded. This is expected, as its removal introduces noise by labeling valid refusals with negative feedback, which induces a bias against refusals. Notably, the Refusal Validation strategy yields only 572 instances, but it has a substantial impact on the final experimental results.
This suggests that the safety boundary of reward models is sensitive, requiring further efforts to enhance its robustness. It further suggests that in-the-wild human interactions contain much subtle noise, requiring a robust pipeline to extract valid human feedback.

\begin{table}[t]
    \centering
    \small
    \resizebox{\linewidth}{!}{
    \begin{tabular}{lrrrrr}
    \toprule
    Model & Chat & Math & Code & SRF & SRP \\
    \midrule
    \ourmodelsmall & $79.3$ & $75.6$ & $65.8$ & $90.4$ & $72.0$\\
    \midrule
    w/o Feedback Mining & $77.3$ & $73.6$ & $65.5$ & $68.3$ & $77.5$ \\
    w/o Refusal Validation & $80.0$ & $74.6$ & $64.8$ & $28.5$ & $97.0$ \\
    w/o All & $77.0$ & $74.5$ & $64.8$ & $39.7$ & $95.5$ \\
    \bottomrule
    \end{tabular}
    }
    \caption{Ablation study results (\%) on RM-Bench Normal. The variants ``w/o Feedback Mining'' and ``w/o Refusal Validation'' mean models trained without the corresponding data. ``SRF'' and ``SRP'' denote Safety-Refusal and Safety-Response subsets, respectively.}
    \label{tab:ablation}
\end{table}

\begin{figure}
    \centering
    \includegraphics[width=0.9\linewidth]{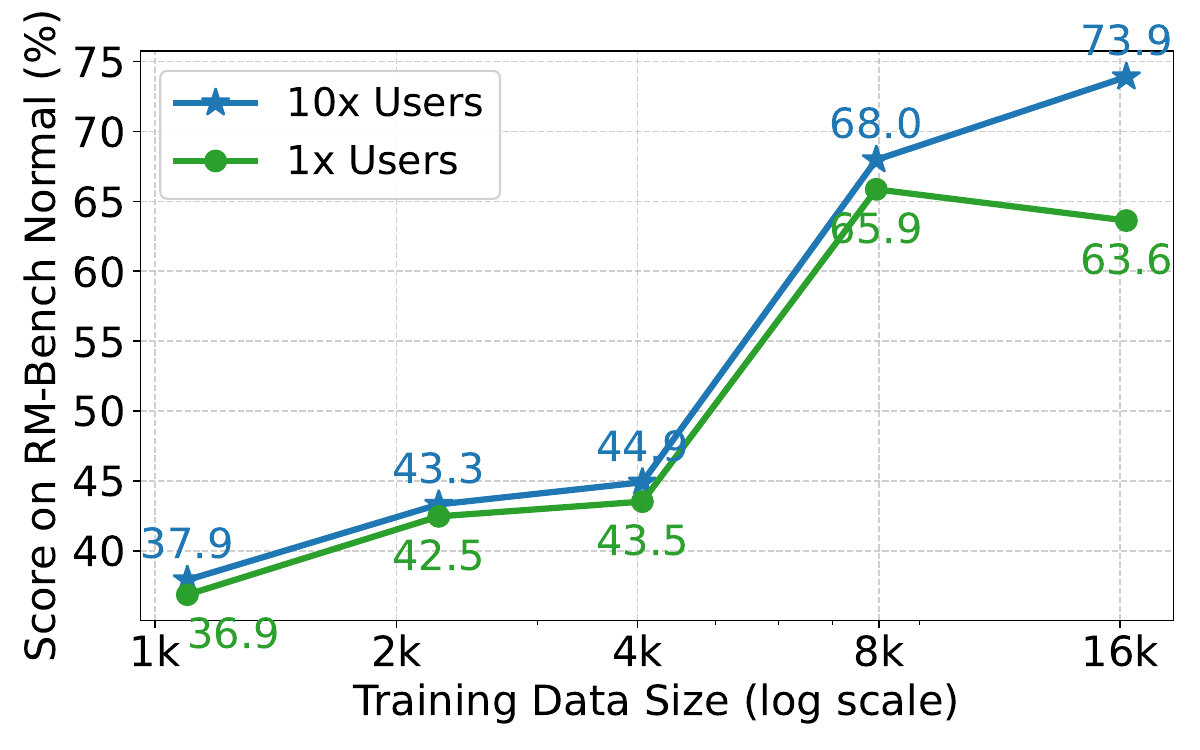}
    \caption{Performance on RM-Bench Normal across varying data sizes and user counts.}
    \label{fig:user_diversity}
\end{figure}

A significant advantage of in-the-wild human interactions lies in the user diversity, rather than a limited group of human annotators. 
To investigate this, we conduct an analysis on the impact of user diversity on training reward models. We measure user diversity by the number of unique users. We construct two series of datasets of identical size, where one contains ten times the number of unique users as the other. The results of trained reward models are shown in Figure~\ref{fig:user_diversity}.
We can observe that: 
(1) Generally, model performance improves with the training data size, which aligns with the data scaling law~\citep{kaplan2020scaling}.
(2) For a given data size, models trained on data from a larger number of users consistently perform better. It demonstrates that the model benefits directly from user diversity, as more diverse users may provide more robust feedback.
In conclusion, scaling up the size and diversity of human interactions is promising for developing advanced reward models.

\begin{table*}
    \centering
    \small
    \resizebox{\linewidth}{!}{
    \begin{tabular}{lrrrrrrrr}
    \toprule
    Model & GSM8K & MATH-500 & MMLU Pro &IFEval & Alpaca Eval 2.0 & Arena Hard & Average \\
    \midrule
    Llama3.1-8B-Instruct & $83.6$ & $49.6$ & $48.0$ &$78.7$ & $33.1$ & $55.7$ & $58.1$ \\
    \midrule
        Offline DPO with \ourmodel & $84.3$ & $48.6$ & $48.5$ & $80.9$ & $33.8$ & $55.4$ & $58.6$\\
    Online DPO with ArmoRM & $86.7$ & $49.4$ & $\mathbf{50.3}$ & $80.6$ & $37.9$ & $57.7$ & $60.4$ \\
    Online DPO with \ourmodel & $\mathbf{87.9}$ & $\mathbf{51.6}$ & $48.9$ & $\mathbf{82.1}$ & $\mathbf{41.3}$ & $\mathbf{63.5}$ & $\mathbf{62.6}$\\
    \bottomrule
    \end{tabular}
    }
    \caption{Results (\%) of the original Llama3.1-8B-Instruct and models trained with different settings. ArmoRM denotes ArmoRM-Llama3-8B-v0.1. We report the length-controlled win rate for Alpaca Eval 2.0.}
    \label{tab:dpo_results}
\end{table*}

\subsection{Analysis on Calibration}
\label{sec:calibration}

A robust reward model should be well-calibrated, meaning that its prediction accuracy should correlate positively with its confidence. 
Here, we focus on binary classification tasks, which is a widely-used setting in reward model benchmarks to select the better of two given responses (chosen and rejected). As detailed in \cref{sec:training}, we employ ordinal regression to train our reward models. A key advantage of this approach is that it directly leverages probabilities as scores, which theoretically promotes good calibration.
In this section, we explore the calibration properties of \ourmodel using the RM-Bench Normal dataset. 
We use the score margin between chosen and rejected responses as a confidence proxy, and apply Platt Scaling~\citep{guo2017calibration} to map raw margins to the $[0, 1]$ range: 
$
\text{confidence}=\sigma(a \times \text{score}_{\text{diff}} + b).
$
Here, $\sigma$ is the sigmoid function, $a$ and $b$ are learnable parameters. We use $50\%$ data to fit $a$ and $b$, and the remaining $50\%$ to evaluate calibration. We adopt the widely used Expected Calibration Error (ECE) as the metric, where a lower value indicates better calibration. \ourmodel achieves a remarkably low ECE of $2.76\%$. This implies that, on average, the discrepancy between the predicted confidence and its actual accuracy is less than $3\%$. 
In comparison, ArmoRM-Llama3-8B-v0.1 yields an ECE of $8.81\%$. It demonstrates that \ourmodel is well-calibrated for pairwise classification tasks.

\begin{figure}
    \centering
    \includegraphics[width=0.9\linewidth]{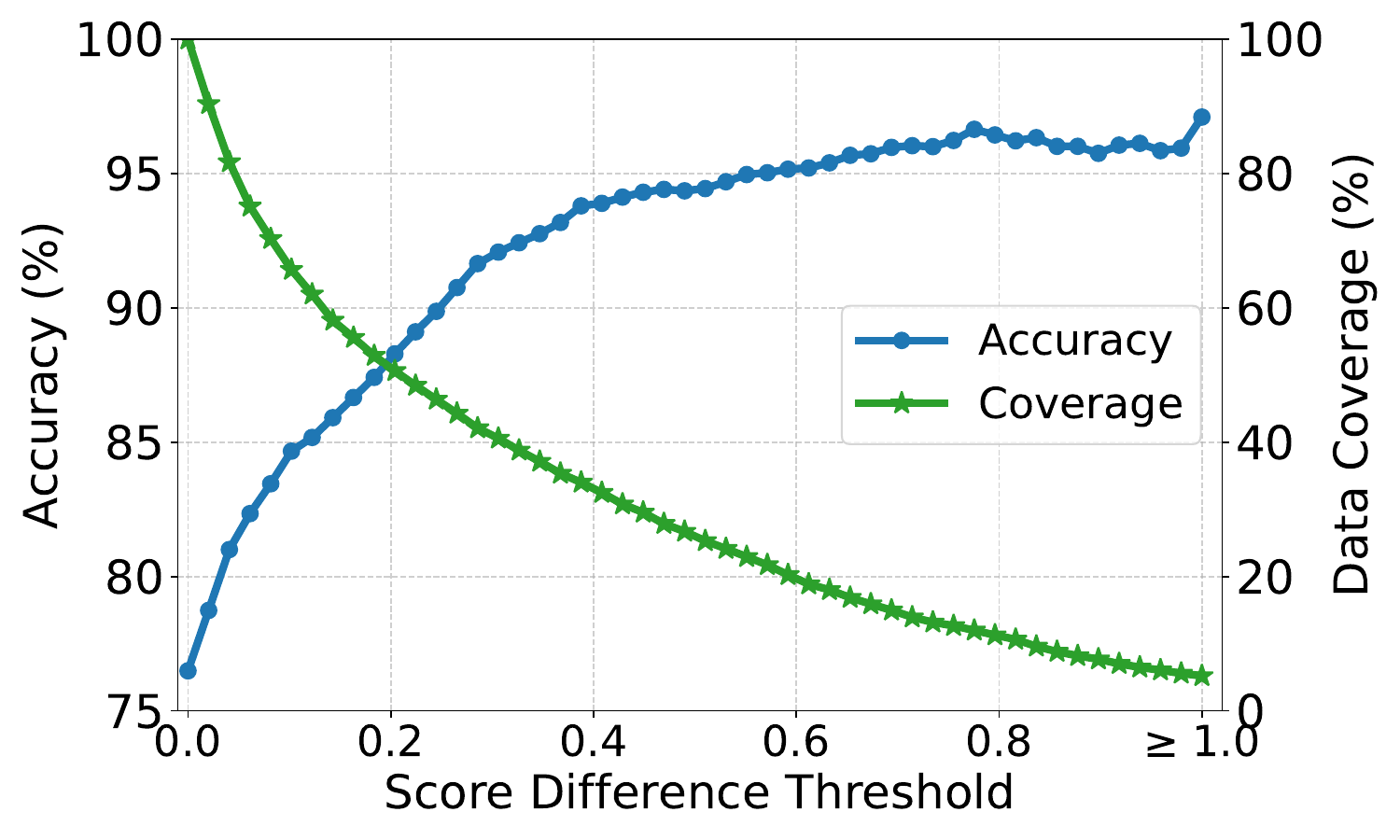}
    \caption{Accuracy and data coverage against score difference threshold, filtering for predictions where the chosen-rejected score margin exceeds the threshold. The results are reported on RM-Bench Normal.}
    \label{fig:threshold_acc}
\end{figure}

We further investigate the accuracy across varying score differences. Specifically, we use a threshold to filter out predictions with score margins below this threshold and re-calculate the accuracy for the remaining subset.
As illustrated in Figure~\ref{fig:threshold_acc}, accuracy consistently improves as the threshold increases. Notably, setting the threshold to $0.2$ improves the accuracy to $87\%$ for retaining about $50\%$ of the predictions. This demonstrates that the score difference serves as a reliable proxy for confidence, allowing for the effective filtering of uncertain predictions. Consequently, \ourmodel can be effectively integrated with stronger LLMs~\citep{xuask25}, external tools~\citep{peng2025agentic}, or even human experts to provide highly precise rewards. We leave this exploration for future work.

\subsection{Analysis on Cross-Sample Consistency}
\label{sec:consistency}

In real-world applications, a robust reward model is expected to exhibit cross-sample consistency, which is to assign absolute scores that are comparable across different queries. This property is essential not only for stabilizing downstream RL training~\citep{xu2025unified} but also for deployment, enabling a unified score threshold to filter out unacceptable responses regardless of the context.
However, existing reward model benchmarks primarily assess local pairwise ranking, determining relative ranking rather than guaranteeing cross-sample consistency.
To address this, we evaluate cross-sample consistency using the ROC-AUC metric on a curated evaluation set from WildChat. The evaluation set is sampled from the WildChat held-out set. Following the pipeline described in \cref{sec:data_construct}, we instead simplify the task from four-category (excluding the \textit{Neutral Ambiguity} category) to binary classification, i.e., positive and negative, to reflect the real user satisfaction (accept or reject) with the response. In this user-centric formulation, we consider the responses with positive feedback as high-quality and those with negative feedback as low-quality. The resulting pointwise evaluation set contains $948$ instances, each consisting of a conversation history, a user query, the corresponding response, and a binary feedback label (positive or negative).
More details are in Appendix~\ref{sec:app_exp_detail}. ROC-AUC measures the probability that a positive instance is ranked higher than a negative one across the global distribution, thereby evaluating cross-sample consistency.

\begin{figure}
    \centering
    \includegraphics[width=0.9\linewidth]{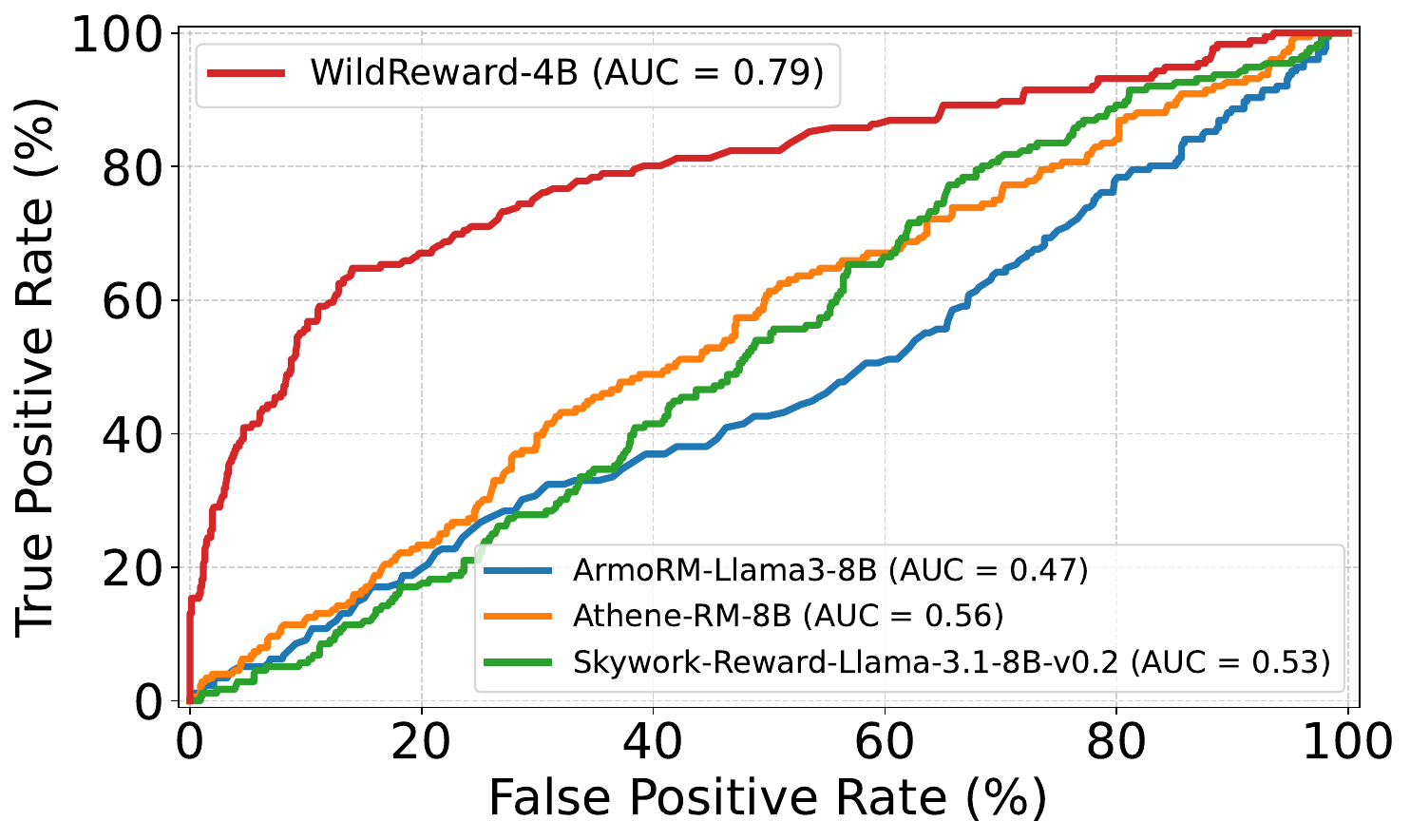}
    \caption{ROC curves and ROC-AUC scores of different reward models in the pointwise evaluation.}
    \label{fig:roc_auc}
\end{figure}

We evaluate \ourmodel and several conventional reward models, and the results are shown in Figure~\ref{fig:roc_auc}. We observe that \ourmodel significantly outperforms the baselines, which are trained on preference pairs using the Bradley-Terry objective. This aligns with the widely recognized limitation that such models often exhibit poor global score calibration~\citep{casper2023open}.
In contrast, \ourmodel is trained via ordinal regression on a global scale, which inherently promotes superior cross-sample consistency.
In conclusion, our analysis demonstrates that mining absolute feedback signals from authentic human interactions offers a promising pathway for global reward calibration.

\subsection{Application in DPO Training}
\label{sec:dpo_training}

Finally, we evaluate the practical utility of reward models by applying them directly to guide policy model training. Specifically, we curate a training dataset of $20,000$ prompts sourced from Infinity Instruct~\citep{li2025infinity} and employ Llama-3.1-8B-Instruct~\citep{dubey2024llama} as the policy model. We conduct Direct Preference Optimization (DPO; \citealp{rafailov2023direct}) using two settings: 
(1) Offline DPO. For each prompt, we first sample four responses offline from the policy model. We then adopt reward models to score these candidates, and select the highest and lowest-scoring responses to form preference pairs for training.
(2) Online DPO. During training, for a batch of prompts, we generate eight responses per prompt on-the-fly using the current policy model. We then adopt reward models to score these responses and select the highest and lowest-scoring responses to form preference pairs.
We evaluate the trained policy model using various widely-used benchmarks, including GSM8K~\citep{cobbe2021training} and MATH-500~\citep{hendrycks2measuring} for mathematical reasoning, MMLU Pro~\citep{wang2024mmlu} for general QA, IFEval~\citep{zhou2023instruction} for instruction following, and Alpaca Eval 2.0~\citep{dubois2024length} and Arena Hard~\citep{licrowdsourced25} for creative writing.
More details are placed in Appendix~\ref{sec:app_exp_detail}.

The experimental results are presented in Table~\ref{tab:dpo_results}. We can observe that: 
(1) Online DPO with \ourmodel yields significant gains over Llama3.1-8B-Instruct and also outperforms ArmoRM. Given that Llama3.1 has already undergone extensive DPO training on large-scale datasets~\citep{dubey2024llama}, the additional improvements demonstrate that \ourmodel provides effective rewards in guiding policy optimization. Furthermore, surpassing ArmoRM suggests that \ourmodel also learn an effective reward scoring mechanism even without training on preference pairs. Due to computational constraints, we adopt only about 20k training prompts. We believe scaling the training data holds potential for even better performance.
(2) Improvements are most significant on Alpaca Eval 2.0 and Arena Hard, which serve as proxies for subjective human evaluation. This demonstrates that \ourmodel captures authentic human preferences. We also observe gains in mathematical reasoning and instruction following, suggesting that \ourmodel also evaluates objective response quality effectively.
(3) Offline DPO yields nearly no improvements. The reason may be that it suffers from a distribution shift, where static data fails to align with the evolving policy distribution~\citep{guo2024direct}. This suggests that online learning is an effective way for further enhancing model capabilities. Consequently, we advocate that future research on reward models adopt online DPO to validate model effectiveness. In conclusion, \ourmodel effectively guides policy online optimization. As an initial step to explore training reward models from human interactions, we believe developing dynamic reward models that coevolve with new human interactions is a promising direction. We leave this exploration for future work.

%



\section{Related Work}



This work primarily focuses on reward modeling for large language models. Since the introduction of Reinforcement Learning from Human Feedback (RLHF; \citealp{ouyang2022training}), which utilizes reward models as proxies for human feedback to train LLMs and enhance their alignment, the standard practice to train reward models has been to collect extensive preference pairs and train models via the Bradley-Terry (BT; \citealp{bradley1952rank}) objective. 
There are numerous studies aiming to develop more advanced reward models, generally focusing on three key directions:
(1) Expanding preference data. Motivated by data scaling laws, this direction of research focuses on automatically or manually collecting larger and more diverse sets of preference pairs to develop more advanced reward models~\citep{bai2022constitutional, lee2024rlaif,park2024offsetbias,cui2024ultrafeedback,zollo2024personalllm,zhu2024starling,wang2024helpsteer,liu2024skywork,liu2025skywork,wang2025helpsteer3}.
(2) Improving objectives and model architectures to enhance robustness and discriminative capability of reward models~\citep{chen2024odin,li2024aligning,yang2024bayesian,wang2024interpretable,liu2025lipo,liurrm25}.
(3) Novel modeling frameworks. This direction involves developing new modeling methods, such as training pairwise reward models~\citep{jiang2023llm, xu2025unified,liu2025pairwise}, generative reward models~\citep{kim2023prometheus,liu2025inference,chen2025rm,guo2025reward}, and hybrid systems integrated with external tools~\citep{li2024tool,liao2025rlmr,zhang2025longreward,peng2025agentic}. Nonetheless, the training of conventional reward models relies heavily on preference pairs.

Therefore, recent efforts have focused on leveraging implicit feedback from massive human-LLM interactions in the wild. A widely used approach is to collect negative feedback from interactions to rewrite responses, constructing preference pairs for DPO training~\citep{shi2024wildfeedback,jin2025era} or directly for SFT training~\citep{liu2025user}. However, these approaches do not train a reward model, relying solely on static human feedback. This limits their scalability and applications. For example, they can not generalize to new queries or support online learning methods such as Online DPO.
Notably, there are two studies by \citet{han2025reinforcement} and \citet{pang2024leveraging} that are closely related to our work. They extract feedback from human interactions to train a binary classifier that predicts whether a user is satisfied with a response. However, this binary classifier fails to capture response ranking information and can not serve as a reward model. Consequently, its application is limited to the Best-of-N search~\citep{pang2024leveraging} or integration with other reward models~\citep{han2025reinforcement}.
In this work, we train an advanced reward model \ourmodel directly from interaction data without using preference pairs. \ourmodel achieves impressive performance with improved calibration.

\section{Conclusion}

In this work, we explore training reward models from in-the-wild human interactions. We leverage WildChat as our interaction source and propose an automated pipeline to extract valid human feedback, resulting in \ourdata, a curated dataset of 186k high-quality instances. Using this dataset, we train \ourmodel directly via ordinal regression without preference pairs. Extensive experiments demonstrate the efficacy of \ourmodel, with improved calibration and cross-sample consistency. 
Given the growing scale of in-the-wild interactions, our work highlights a promising direction for leveraging these valuable resources, and we encourage more efforts to explore this area in the future.


\section*{Limitations}

This section discusses the limitations of our work, which are primarily threefold:
(1) Regarding the dataset \ourdata, this work only considers English and Chinese conversations within the WildChat dataset. This may limit the broader application of our reward models to other languages. We encourage the research community to use more languages to develop more advanced and multilingual reward models.
(2) Regarding \ourmodel, we do not perform a sufficient search for optimal configurations, e.g., hyper-parameters or backbone models, when training our reward models. Other settings could yield superior performance. Due to the computational constraints, we leave a more thorough exploration of the configuration space to future work.
(3) Regarding applications in policy training, we do not employ \ourmodel for RL training. Because it is highly resource-intensive and non-trivial to stabilize the RL training. We adopt online DPO training using \ourmodel and the results demonstrate the effectiveness of our model.

\section*{Ethical Considerations}
We discuss potential ethical concerns as follows:
(1) Intellectual property. This work mainly uses two open-sourced datasets, WildChat and Infinity-Instruct. WildChat is released under the ODC-By license\footnote{\url{https://opendatacommons.org/licenses/by/1-0/}}. Infinity-Instruct is released under the CC BY-NC 4.0 license\footnote{\url{https://creativecommons.org/licenses/by-nc/4.0/}}. We strictly adhere to their licenses and terms of use. Our dataset \ourdata will be released under Apache License 2.0\footnote{\url{https://www.apache.org/licenses/LICENSE-2.0}}.
(2) Intended use. \ourdata consists of authentic human feedback and is designed for training reward models. From this dataset, we train \ourmodel, a reward model that scores the quality of generated responses, which can be used for training or test-time scaling.
(3) Potential risk control. \ourdata is derived from WildChat. We believe it is properly anonymized. We do not introduce any additional sensitive information. \ourmodel is trained on real-world data and may inherently contain biases. Users should not exploit these biases for malicious purposes, such as reward hacking. We advise that the usage of \ourmodel should undergo a verification process before usage.
(4) AI assistance. We adopt Gemini to polish some sentences.

\bibliography{custom}

@article{ouyang2022training,
  title={Training language models to follow instructions with human feedback},
  author={Ouyang, Long and Wu, Jeffrey and Jiang, Xu and Almeida, Diogo and Wainwright, Carroll and Mishkin, Pamela and Zhang, Chong and Agarwal, Sandhini and Slama, Katarina and Ray, Alex and others},
  journal={Advances in neural information processing systems},
  volume={35},
  pages={27730--27744},
  year={2022}
}

@article{wang2025helpsteer3,
  title={HelpSteer3-Preference: Open Human-Annotated Preference Data across Diverse Tasks and Languages},
  author={Wang, Zhilin and Zeng, Jiaqi and Delalleau, Olivier and Shin, Hoo-Chang and Soares, Felipe and Bukharin, Alexander and Evans, Ellie and Dong, Yi and Kuchaiev, Oleksii},
  journal={arXiv preprint arXiv:2505.11475},
  year={2025}
}

@article{wang2024helpsteer,
  title={Helpsteer 2: Open-source dataset for training top-performing reward models},
  author={Wang, Zhilin and Dong, Yi and Delalleau, Olivier and Zeng, Jiaqi and Shen, Gerald and Egert, Daniel and Zhang, Jimmy and Sreedhar, Makesh Narsimhan and Kuchaiev, Oleksii},
  journal={Advances in Neural Information Processing Systems},
  volume={37},
  pages={1474--1501},
  year={2024}
}

@article{liu2025skywork,
  title={Skywork-Reward-V2: Scaling Preference Data Curation via Human-AI Synergy},
  author={Liu, Chris Yuhao and Zeng, Liang and Xiao, Yuzhen and He, Jujie and Liu, Jiacai and Wang, Chaojie and Yan, Rui and Shen, Wei and Zhang, Fuxiang and Xu, Jiacheng and others},
  journal={arXiv preprint arXiv:2507.01352},
  year={2025}
}

@inproceedings{zhaowildchat24,
  title={WildChat: 1M ChatGPT Interaction Logs in the Wild},
  author={Zhao, Wenting and Ren, Xiang and Hessel, Jack and Cardie, Claire and Choi, Yejin and Deng, Yuntian},
  booktitle={Proceedings of ICLR},
year=2024
}

@inproceedings{zhenglmsys24,
  title={LMSYS-Chat-1M: A Large-Scale Real-World LLM Conversation Dataset},
  author={Zheng, Lianmin and Chiang, Wei-Lin and Sheng, Ying and Li, Tianle and Zhuang, Siyuan and Wu, Zhanghao and Zhuang, Yonghao and Li, Zhuohan and Lin, Zi and Xing, Eric and others},
  booktitle={Proceedings of ICLR},
year=2024
}

@article{wang2025survey,
  title={A Survey on Ordinal Regression: Applications, Advances and Prospects},
  author={Wang, Jinhong and Chen, Jintai and Liu, Jian and Tang, Dongqi and Chen, Danny Z and Wu, Jian},
  journal={arXiv preprint arXiv:2503.00952},
  year={2025}
}

@article{agarwal2025gpt,
  title={gpt-oss-120b \& gpt-oss-20b model card},
  author={Agarwal, Sandhini and Ahmad, Lama and Ai, Jason and Altman, Sam and Applebaum, Andy and Arbus, Edwin and Arora, Rahul K and Bai, Yu and Baker, Bowen and Bao, Haiming and others},
  journal={arXiv preprint arXiv:2508.10925},
  year={2025}
}

@inproceedings{lambert2025rewardbench,
  title={Rewardbench: Evaluating reward models for language modeling},
  author={Lambert, Nathan and Pyatkin, Valentina and Morrison, Jacob and Miranda, Lester James Validad and Lin, Bill Yuchen and Chandu, Khyathi and Dziri, Nouha and Kumar, Sachin and Zick, Tom and Choi, Yejin and others},
  booktitle={Findings of NAACL},
  pages={1755--1797},
  year={2025}
}

@inproceedings{liurm25,
  title={RM-Bench: Benchmarking Reward Models of Language Models with Subtlety and Style},
  author={Liu, Yantao and Yao, Zijun and Min, Rui and Cao, Yixin and Hou, Lei and Li, Juanzi},
  booktitle={Proceedings of ICLR},
year={2025}
}

@inproceedings{frickevaluate25,
  title={How to Evaluate Reward Models for RLHF},
  author={Frick, Evan and Li, Tianle and Chen, Connor and Chiang, Wei-Lin and Angelopoulos, Anastasios Nikolas and Jiao, Jiantao and Zhu, Banghua and Gonzalez, Joseph E and Stoica, Ion},
  booktitle={Proceedings of ICLR},
year={2025}
}

@inproceedings{tanjudgebench25,
  title={JudgeBench: A Benchmark for Evaluating LLM-Based Judges},
  author={Tan, Sijun and Zhuang, Siyuan and Montgomery, Kyle and Tang, William Yuan and Cuadron, Alejandro and Wang, Chenguang and Popa, Raluca and Stoica, Ion},
  booktitle={Proceedings of ICLR},
year={2025}
}

@inproceedings{xuask25,
  title={Ask a Strong LLM Judge when Your Reward Model is Uncertain},
  author={Xu, Zhenghao and Lu, Qin and Zhang, Qingru and Qiu, Liang and Hong, Ilgee and Yu, Changlong and Yao, Wenlin and Liu, Yao and Jiang, Haoming and Li, Lihong and others},
  booktitle={The Thirty-ninth Annual Conference on Neural Information Processing Systems},
year={2025}
}

@article{rafailov2023direct,
  title={Direct preference optimization: Your language model is secretly a reward model},
  author={Rafailov, Rafael and Sharma, Archit and Mitchell, Eric and Manning, Christopher D and Ermon, Stefano and Finn, Chelsea},
  journal={Advances in neural information processing systems},
  volume={36},
  pages={53728--53741},
  year={2023}
}

@article{bradley1952rank,
  title={Rank analysis of incomplete block designs: I. the method of paired comparisons},
  author={Bradley, Ralph Allan and Terry, Milton E},
  journal={Biometrika},
  volume={39},
  number={3/4},
  pages={324--345},
  year={1952},
  publisher={JSTOR}
}

@article{yang2025qwen3,
  title={Qwen3 technical report},
  author={Yang, An and Li, Anfeng and Yang, Baosong and Zhang, Beichen and Hui, Binyuan and Zheng, Bo and Yu, Bowen and Gao, Chang and Huang, Chengen and Lv, Chenxu and others},
  journal={arXiv preprint arXiv:2505.09388},
  year={2025}
}

@inproceedings{park2024offsetbias,
  title={Offsetbias: Leveraging debiased data for tuning evaluators},
  author={Park, Junsoo and Jwa, Seungyeon and Meiying, Ren and Kim, Daeyoung and Choi, Sanghyuk},
  booktitle={Findings of EMNLP},
  pages={1043--1067},
  year={2024}
}

@inproceedings{wang2024interpretable,
  title={Interpretable Preferences via Multi-Objective Reward Modeling and Mixture-of-Experts},
  author={Wang, Haoxiang and Xiong, Wei and Xie, Tengyang and Zhao, Han and Zhang, Tong},
  booktitle={Findings of EMNLP},
  pages={10582--10592},
  year={2024}
}

@misc{Athene2024,
    title = {Athene-70B: Redefining the Boundaries of Post-Training for Open Models},
    url = {https://nexusflow.ai/blogs/athene},
    author = {Frick, Evan and Jin, Peter and Li, Tianle and Ganesan, Karthik and Zhang, Jian and Jiao, Jiantao and Zhu, Banghua},    
    month = {July},
    year = {2024}
}

@article{liu2024skywork,
  title={Skywork-reward: Bag of tricks for reward modeling in llms},
  author={Liu, Chris Yuhao and Zeng, Liang and Liu, Jiacai and Yan, Rui and He, Jujie and Wang, Chaojie and Yan, Shuicheng and Liu, Yang and Zhou, Yahui},
  journal={arXiv preprint arXiv:2410.18451},
  year={2024}
}

@article{cai2024internlm2,
  title={Internlm2 technical report},
  author={Cai, Zheng and Cao, Maosong and Chen, Haojiong and Chen, Kai and Chen, Keyu and Chen, Xin and Chen, Xun and Chen, Zehui and Chen, Zhi and Chu, Pei and others},
  journal={arXiv preprint arXiv:2403.17297},
  year={2024}
}

@misc{INF-ORM-Llama3.1-70B, 
      url={[https://huggingface.co/infly/INF-ORM-Llama3.1-70B](https://huggingface.co/infly/INF-ORM-Llama3.1-70B)},
      title={INF-ORM-Llama3.1-70B},
      year={2024},
      author={Minghao Yang, Chao Qu, Xiaoyu Tan}
}

@misc{wang2024helpsteer2preferencecomplementingratingspreferences,
      title={HelpSteer2-Preference: Complementing Ratings with Preferences}, 
      author={Zhilin Wang and Alexander Bukharin and Olivier Delalleau and Daniel Egert and Gerald Shen and Jiaqi Zeng and Oleksii Kuchaiev and Yi Dong},
      year={2024},
      eprint={2410.01257},
      archivePrefix={arXiv},
      primaryClass={cs.LG},
      url={https://arxiv.org/abs/2410.01257}, 
}

@article{kaplan2020scaling,
  title={Scaling laws for neural language models},
  author={Kaplan, Jared and McCandlish, Sam and Henighan, Tom and Brown, Tom B and Chess, Benjamin and Child, Rewon and Gray, Scott and Radford, Alec and Wu, Jeffrey and Amodei, Dario},
  journal={arXiv preprint arXiv:2001.08361},
  year={2020}
}

@inproceedings{guo2017calibration,
  title={On calibration of modern neural networks},
  author={Guo, Chuan and Pleiss, Geoff and Sun, Yu and Weinberger, Kilian Q},
  booktitle={International conference on machine learning},
  pages={1321--1330},
  year={2017},
  organization={PMLR}
}

@article{peng2025agentic,
  title={Agentic reward modeling: Integrating human preferences with verifiable correctness signals for reliable reward systems},
  author={Peng, Hao and Qi, Yunjia and Wang, Xiaozhi and Yao, Zijun and Xu, Bin and Hou, Lei and Li, Juanzi},
  journal={arXiv preprint arXiv:2502.19328},
  year={2025}
}

@article{dubois2024length,
  title={Length-controlled alpacaeval: A simple way to debias automatic evaluators},
  author={Dubois, Yann and Galambosi, Bal{\'a}zs and Liang, Percy and Hashimoto, Tatsunori B},
  journal={arXiv preprint arXiv:2404.04475},
  year={2024}
}

@article{xu2025unified,
  title={A Unified Pairwise Framework for RLHF: Bridging Generative Reward Modeling and Policy Optimization},
  author={Xu, Wenyuan and Zuo, Xiaochen and Xin, Chao and Yue, Yu and Yan, Lin and Wu, Yonghui},
  journal={arXiv preprint arXiv:2504.04950},
  year={2025}
}

@article{casper2023open,
  title={Open Problems and Fundamental Limitations of Reinforcement Learning from Human Feedback},
  author={Casper, Stephen and Davies, Xander and Shi, Claudia and Krendl Gilbert, Thomas and Scheurer, J{\'e}r{\'e}my and Rando Ramirez, Javier and Freedman, Rachel and Korbak, Tomasz and Lindner, David and Freire, Pedro and others},
  journal={Transactions on Machine Learning Research},
  year={2023}
}

@article{li2025infinity,
  title={Infinity Instruct: Scaling Instruction Selection and Synthesis to Enhance Language Models},
  author={Li, Jijie and Du, Li and Zhao, Hanyu and Zhang, Bo-wen and Wang, Liangdong and Gao, Boyan and Liu, Guang and Lin, Yonghua},
  journal={arXiv preprint arXiv:2506.11116},
  year={2025}
}

@article{dubey2024llama,
  title={The llama 3 herd of models},
  author={Dubey, Abhimanyu and Jauhri, Abhinav and Pandey, Abhinav and Kadian, Abhishek and Al-Dahle, Ahmad and Letman, Aiesha and Mathur, Akhil and Schelten, Alan and Yang, Amy and Fan, Angela and others},
  journal={arXiv preprint arXiv:2407.21783},
  year={2024}
}

@article{cobbe2021training,
  title={Training verifiers to solve math word problems},
  author={Cobbe, Karl and Kosaraju, Vineet and Bavarian, Mohammad and Chen, Mark and Jun, Heewoo and Kaiser, Lukasz and Plappert, Matthias and Tworek, Jerry and Hilton, Jacob and Nakano, Reiichiro and others},
  journal={arXiv preprint arXiv:2110.14168},
  year={2021}
}

@inproceedings{hendrycks2measuring,
  title={Measuring Mathematical Problem Solving With the MATH Dataset},
  author={Hendrycks, Dan and Burns, Collin and Kadavath, Saurav and Arora, Akul and Basart, Steven and Tang, Eric and Song, Dawn and Steinhardt, Jacob},
  booktitle={Thirty-fifth Conference on Neural Information Processing Systems Datasets and Benchmarks Track (Round 2)},
year={2021}
}

@article{wang2024mmlu,
  title={Mmlu-pro: A more robust and challenging multi-task language understanding benchmark},
  author={Wang, Yubo and Ma, Xueguang and Zhang, Ge and Ni, Yuansheng and Chandra, Abhranil and Guo, Shiguang and Ren, Weiming and Arulraj, Aaran and He, Xuan and Jiang, Ziyan and others},
  journal={Advances in Neural Information Processing Systems},
  volume={37},
  pages={95266--95290},
  year={2024}
}

@article{zhou2023instruction,
  title={Instruction-following evaluation for large language models},
  author={Zhou, Jeffrey and Lu, Tianjian and Mishra, Swaroop and Brahma, Siddhartha and Basu, Sujoy and Luan, Yi and Zhou, Denny and Hou, Le},
  journal={arXiv preprint arXiv:2311.07911},
  year={2023}
}

@inproceedings{licrowdsourced25,
  title={From Crowdsourced Data to High-quality Benchmarks: Arena-Hard and Benchbuilder Pipeline},
  author={Li, Tianle and Chiang, Wei-Lin and Frick, Evan and Dunlap, Lisa and Wu, Tianhao and Zhu, Banghua and Gonzalez, Joseph E and Stoica, Ion},
  booktitle={Proceedings of ICML},
year={2025}
}

@article{guo2024direct,
  title={Direct language model alignment from online ai feedback},
  author={Guo, Shangmin and Zhang, Biao and Liu, Tianlin and Liu, Tianqi and Khalman, Misha and Llinares, Felipe and Rame, Alexandre and Mesnard, Thomas and Zhao, Yao and Piot, Bilal and others},
  journal={arXiv preprint arXiv:2402.04792},
  year={2024}
}

@article{bai2022constitutional,
  title={Constitutional ai: Harmlessness from ai feedback},
  author={Bai, Yuntao and Kadavath, Saurav and Kundu, Sandipan and Askell, Amanda and Kernion, Jackson and Jones, Andy and Chen, Anna and Goldie, Anna and Mirhoseini, Azalia and McKinnon, Cameron and others},
  journal={arXiv preprint arXiv:2212.08073},
  year={2022}
}

@inproceedings{lee2024rlaif,
  title={RLAIF vs. RLHF: Scaling Reinforcement Learning from Human Feedback with AI Feedback},
  author={Lee, Harrison and Phatale, Samrat and Mansoor, Hassan and Mesnard, Thomas and Ferret, Johan and Lu, Kellie Ren and Bishop, Colton and Hall, Ethan and Carbune, Victor and Rastogi, Abhinav and others},
  booktitle={International Conference on Machine Learning},
  pages={26874--26901},
  year={2024},
  organization={PMLR}
}

@inproceedings{cui2024ultrafeedback,
  title={ULTRAFEEDBACK: Boosting Language Models with Scaled AI Feedback},
  author={Cui, Ganqu and Yuan, Lifan and Ding, Ning and Yao, Guanming and He, Bingxiang and Zhu, Wei and Ni, Yuan and Xie, Guotong and Xie, Ruobing and Lin, Yankai and others},
  booktitle={Proceedings of ICML},
  year={2024}
}

@inproceedings{zhu2024starling,
  title={Starling-7b: Improving helpfulness and harmlessness with rlaif},
  author={Zhu, Banghua and Frick, Evan and Wu, Tianhao and Zhu, Hanlin and Ganesan, Karthik and Chiang, Wei-Lin and Zhang, Jian and Jiao, Jiantao},
  booktitle={First Conference on Language Modeling},
  year={2024}
}

@article{zollo2024personalllm,
  title={Personalllm: Tailoring llms to individual preferences},
  author={Zollo, Thomas P and Siah, Andrew Wei Tung and Ye, Naimeng and Li, Ang and Namkoong, Hongseok},
  journal={arXiv preprint arXiv:2409.20296},
  year={2024}
}

@inproceedings{chen2024odin,
  title={ODIN: Disentangled Reward Mitigates Hacking in RLHF},
  author={Chen, Lichang and Zhu, Chen and Chen, Jiuhai and Soselia, Davit and Zhou, Tianyi and Goldstein, Tom and Huang, Heng and Shoeybi, Mohammad and Catanzaro, Bryan},
  booktitle={International Conference on Machine Learning},
  pages={7935--7952},
  year={2024},
  organization={PMLR}
}

@inproceedings{yang2024bayesian,
  title={Bayesian Reward Models for LLM Alignment},
  author={Yang, Adam X and Robeyns, Maxime and Coste, Thomas and Shi, Zhengyan and Wang, Jun and Ammar, Haitham Bou and Aitchison, Laurence},
  booktitle={ICML 2024 Workshop on Structured Probabilistic Inference $\{$$\backslash$\&$\}$ Generative Modeling},
year={2024}
}

@inproceedings{li2024aligning,
  title={Aligning Crowd Feedback via Distributional Preference Reward Modeling},
  author={Li, Dexun and Zhang, Cong and Dong, Kuicai and Deik, Derrick Goh Xin and Tang, Ruiming and Liu, Yong},
  booktitle={ICML 2024 Workshop on Models of Human Feedback for AI Alignment},
year={2024}
}

@inproceedings{liurrm25,
  title={RRM: Robust Reward Model Training Mitigates Reward Hacking},
  author={Liu, Tianqi and Xiong, Wei and Ren, Jie and Chen, Lichang and Wu, Junru and Joshi, Rishabh and Gao, Yang and Shen, Jiaming and Qin, Zhen and Yu, Tianhe and others},
  booktitle={The Thirteenth International Conference on Learning Representations},
year={2025}
}

@inproceedings{liu2025lipo,
  title={Lipo: Listwise preference optimization through learning-to-rank},
  author={Liu, Tianqi and Qin, Zhen and Wu, Junru and Shen, Jiaming and Khalman, Misha and Joshi, Rishabh and Zhao, Yao and Saleh, Mohammad and Baumgartner, Simon and Liu, Jialu and others},
  booktitle={Proceedings of ACL},
  year={2025}
}

@inproceedings{jiang2023llm,
  title={LLM-Blender: Ensembling Large Language Models with Pairwise Ranking and Generative Fusion},
  author={Jiang, Dongfu and Ren, Xiang and Lin, Bill Yuchen},
  booktitle={Proceedings of ACL},
  pages={14165--14178},
  year={2023}
}

@article{liu2025pairwise,
  title={Pairwise rm: Perform best-of-n sampling with knockout tournament},
  author={Liu, Yantao and Yao, Zijun and Min, Rui and Cao, Yixin and Hou, Lei and Li, Juanzi},
  journal={arXiv e-prints},
  pages={arXiv--2501},
  year={2025}
}

@inproceedings{kim2023prometheus,
  title={Prometheus: Inducing fine-grained evaluation capability in language models},
  author={Kim, Seungone and Shin, Jamin and Cho, Yejin and Jang, Joel and Longpre, Shayne and Lee, Hwaran and Yun, Sangdoo and Shin, Seongjin and Kim, Sungdong and Thorne, James and others},
  booktitle={The Twelfth International Conference on Learning Representations},
  year={2023}
}

@article{liu2025inference,
  title={Inference-time scaling for generalist reward modeling},
  author={Liu, Zijun and Wang, Peiyi and Xu, Runxin and Ma, Shirong and Ruan, Chong and Li, Peng and Liu, Yang and Wu, Yu},
  journal={arXiv preprint arXiv:2504.02495},
  year={2025}
}

@article{chen2025rm,
  title={Rm-r1: Reward modeling as reasoning},
  author={Chen, Xiusi and Li, Gaotang and Wang, Ziqi and Jin, Bowen and Qian, Cheng and Wang, Yu and Wang, Hongru and Zhang, Yu and Zhang, Denghui and Zhang, Tong and others},
  journal={arXiv preprint arXiv:2505.02387},
  year={2025}
}

@article{guo2025reward,
  title={Reward reasoning model},
  author={Guo, Jiaxin and Chi, Zewen and Dong, Li and Dong, Qingxiu and Wu, Xun and Huang, Shaohan and Wei, Furu},
  journal={arXiv preprint arXiv:2505.14674},
  year={2025}
}

@inproceedings{zhang2025longreward,
  title={Longreward: Improving long-context large language models with ai feedback},
  author={Zhang, Jiajie and Hou, Zhongni and Lv, Xin and Cao, Shulin and Hou, Zhenyu and Niu, Yilin and Hou, Lei and Dong, Yuxiao and Feng, Ling and Li, Juanzi},
  booktitle={Proceedings of ACL},
  pages={3718--3739},
  year={2025}
}

@inproceedings{li2024tool,
  title={Tool-Augmented Reward Modeling},
  author={Li, Lei and Chai, Yekun and Wang, Shuohuan and Sun, Yu and Tian, Hao and Zhang, Ningyu and Wu, Hua},
  booktitle={Proceedings of ICLR},
  year={2024}
}

@article{liao2025rlmr,
  title={Rlmr: Reinforcement learning with mixed rewards for creative writing},
  author={Liao, Jianxing and Zhang, Tian and Feng, Xiao and Zhang, Yusong and Yang, Rui and Wang, Haorui and Wen, Bosi and Wang, Ziying and Shi, Runzhi},
  journal={arXiv preprint arXiv:2508.18642},
  year={2025}
}

@inproceedings{liu2025user,
  title={User feedback in human-LLM dialogues: a lens to understand users but noisy as a learning signal},
  author={Liu, Yuhan and Zhang, Michael JQ and Choi, Eunsol},
  booktitle={Proceedings of EMNLP},
  pages={2666--2681},
  year={2025}
}

@article{jin2025era,
  title={The era of real-world human interaction: Rl from user conversations},
  author={Jin, Chuanyang and Xu, Jing and Liu, Bo and Tao, Leitian and Golovneva, Olga and Shu, Tianmin and Zhao, Wenting and Li, Xian and Weston, Jason},
  journal={arXiv preprint arXiv:2509.25137},
  year={2025}
}

@inproceedings{shi2024wildfeedback,
  title={WildFeedback: Aligning LLMs With In-situ User Interactions And Feedback},
  author={Shi, Taiwei and Wang, Zhuoer and Yang, Longqi and Lin, Ying-Chun and He, Zexue and Wan, Mengting and Zhou, Pei and Jauhar, Sujay Kumar and Xu, Xiaofeng and Song, Xia and others},
  booktitle={NeurIPS 2024 Workshop on Behavioral Machine Learning},
year={2024}
}

@article{han2025reinforcement,
  title={Reinforcement Learning from User Feedback},
  author={Han, Eric and Chen, Jun and Sankararaman, Karthik Abinav and Peng, Xiaoliang and Xu, Tengyu and Helenowski, Eryk and Peng, Kaiyan and Kumar, Mrinal and Wang, Sinong and Fang, Han and others},
  journal={arXiv preprint arXiv:2505.14946},
  year={2025}
}

@inproceedings{pang2024leveraging,
  title={Leveraging implicit feedback from deployment data in dialogue},
  author={Pang, Richard Yuanzhe and Roller, Stephen and Cho, Kyunghyun and He, He and Weston, Jason},
  booktitle={Proceedings of EACL},
  pages={60--75},
  year={2024}
}
\appendix

\clearpage
\section{\ourdata Construction Details}
\label{sec:app_data_construct}

We adopt the WildChat-4.8M\footnote{\url{https://huggingface.co/datasets/allenai/WildChat-4.8M}} dataset as a source of real-world human-LLM interactions. Given the much noise in real conversations, we implement a heavy filtering strategy to curate the dataset. The specific filtering criteria are as follows:

1. Language restriction. We retain only English and Chinese conversations, as these are the primary languages of focus for our study.

2. Exclusion of multimodal queries. Queries that require multimodal understanding or generation, such as image processing, are removed.

3. Exclusion of tool-dependent queries.
We filter out any conversations needing the use of external tools, such as web searches or running Python code.

4. Exclusion of trivial and identity queries. We filter out trivial queries, such as those asking about the model's identity, e.g., ``Who are you?''.

5. Exclusion of context-dependent queries. We remove queries that rely on external context, such as ``Read the information from this document''.

6. Length-based filtering. We filter out queries with conversation histories exceeding 20 turns, as they likely contain much irrelevant information. We also exclude queries with fewer than five words or responses shorter than ten words.

We manually write a series of rules and regular expressions of the above criteria for efficient filtering. 
After the filtering process, as mentioned in \cref{sec:pre_analysis}, we sample $10,000$ instances and employ gpt-oss-120b to classify user follow-up queries into three feedback categories: \textit{Negative}, \textit{Neutral}, and \textit{Positive}. The prompt used for this task is detailed in Figure~\ref{fig:app_analysis_prompt}.
For the automated feedback mining pipeline detailed in \cref{sec:data_construct}, we adopt gpt-oss-120 to identify and classify implicit user feedback into five distinct categories: \textit{Explicit Rejection}, \textit{Error Correction}, \textit{Neutral Ambiguity}, \textit{Positive Engagement}, and \textit{Explicit Satisfaction}. To minimize annotation noise, we adopt a conservative strategy that defaults to \textit{Neutral Ambiguity} in the absence of strong evidence. The specific prompt for this mining process is shown in Figure~\ref{fig:app_pipeline_prompt}. The prompt for the Refusal Validation step is shown in Figure~\ref{fig:app_refual}.

\begin{figure*}[t]
\begin{tcolorbox}[
    listing only,              
    colback=black!5,
    colframe=black!75,
    arc=2mm,
    boxrule=0.5pt,
    width=\textwidth,
    left=4mm, right=4mm, top=2mm, bottom=2mm,
    listing options={
        basicstyle=\ttfamily\small,
        breaklines=true,        
        showstringspaces=false, 
        columns=flexible,
        escapechar=|,
    },
]
You are an expert annotator. Your task is to infer how satisfied the user was 
with the assistant’s previous response, based solely on the user’s latest message.
\par
\textbf{[IMPORTANT RULES]}

1. Only use strong and explicit evidence from the user's message to classify their satisfaction.

2. Do NOT assume the user is satisfied just because they continue the topic.

3. A follow-up question without clear positive or negative cues should be considered Neutral.

\textbf{[INPUT INFO]}

<User's previous message>: \{prev query\}
\par
<Assistant's previous response>: \{prev response\}
\par
<User's latest message>: \{query\}
\par
Based on the user's latest message, classify their preference toward the assistant's previous response into one of the following categories:
\par

\textbf{[CATEGORIES]} (strong evidence only)

[[1]] NEGATIVE

The user explicitly criticizes, rejects, expresses frustration, or points out a mistake, 
missing constraint, or error in the assistant's response.

Examples: “This is wrong.”, “You didn’t answer my question.”, ``I asked for Python, not C++.''

[[2]] NEUTRAL

The user shows no clear positive or negative attitude. The message is a generic follow-up, 
an unrelated question, or ambiguous.

Examples: “Okay, next question.”, “What is the formula for X?”, “How does this apply to Y?”

[[3]] POSITIVE

The user expresses clear gratitude, satisfaction, or positively builds upon the response 
with explicit approval or interest. This requires a clear positive signal.

Examples: “Thanks, this solves it.”, “Perfect answer.”, “Interesting, what happens if we scale it?”

\textbf{[OUTPUT FORMAT]}

[[<category number>]] <brief reasoning>
\end{tcolorbox}
\caption{The prompt used for the three-class user feedback annotation task in \cref{sec:pre_analysis}.}
\label{fig:app_analysis_prompt}
\end{figure*}

\begin{figure*}[t]
\begin{tcolorbox}[
    listing only,             
    colback=black!5,
    colframe=black!75,
    arc=2mm,
    boxrule=0.5pt,
    width=\textwidth,
    left=4mm, right=4mm, top=2mm, bottom=2mm,
    listing options={
        basicstyle=\ttfamily\small,
        breaklines=true,        
        showstringspaces=false, 
        columns=flexible,
        escapechar=|,
    },
]
You are an expert annotator. Your task is to infer how satisfied the user was 
with the assistant’s previous response, based solely on the user’s latest message.

\textbf{[IMPORTANT RULES]}

1. Only use strong and explicit evidence to classify satisfaction.

2. Do NOT assume the user is satisfied or inspired just because they continue the topic.

3. Users often ask follow-up questions even when they are dissatisfied.

4. Neutral, ambiguous, or topic-extending queries should NOT be labeled as "inspired".

\textbf{[INPUT INFO]}

<User's previous message>: \{prev query\}

<Assistant's previous response>: \{prev response\}

<User's latest message>: \{query\}

Based on the user's latest message, classify their preference toward the assistant's previous response into one of the following categories:

\textbf{[CATEGORIES]} (strong evidence only)

[[1]] CLEARLY NEGATIVE / REJECTION  

User explicitly criticizes, rejects, or expresses frustration. 

Examples: “This is wrong.”, “You didn’t answer my question.”, “No, that’s not what I need.”

[[2]] CORRECTION / ERROR POINTER (Negative)

User points out a mistake, missing constraint, or hallucination in the previous response.
The assistant failed to follow the original instruction perfectly.

Examples: ``You calculated the last step wrong.'', ``I asked for Python, not C++.'', ``You forgot to mention the limitations.'', ``This code doesn't run.''

[[3]] NEUTRAL / UNCLEAR

User shows no clear positive/negative attitude.  
Question is unrelated, generic, or ambiguous.  
May simply continue asking questions without emotional signals.
Examples: “Okay, next question.”, “What is the formula for X?”, “How does this apply to Y?” (no emotional cue). 
IF THE MESSAGE IS UNCLEAR (CATEGORY 3):
1. Optionally, check the assistant response quality. If the response is objectively high-quality (correct, helpful, structured), mark as [[4]].
2. If average, leave as [[3]].
3. If poor (misleading, wrong, unsafe), mark as [[1]].
4. DO NOT mark as [[5]] (that requires explicit user satisfaction).
Examples: “Okay, next question.” → [[3]] or fallback [[4]]/[[1]] based on response quality.

[[4]] POSITIVE ENGAGEMENT (strong evidence only)

User explicitly builds upon the response with positive emotional framing  
(e.g., excitement, interest, approval). Not just continuing the topic.

Examples: “Interesting, then what happens if we scale it?”, “That’s helpful — can we apply it to drones?”, “Great point! What about the dynamic case?”
DO NOT label as 4 unless there is clear positive emotion + meaningful extension.

[[5]] CLEAR SATISFACTION

User expresses gratitude, satisfaction, or says the problem is solved.

Examples: “Thanks, this solves it.”, “Perfect answer.”, “That’s all I need.”

\textbf{[OUTPUT FORMAT]}

[[<category number>]] <brief reasoning>

\end{tcolorbox}
\caption{The prompt used for the five-class user feedback classification used in the pipeline in \cref{sec:data_construct}.}
\label{fig:app_pipeline_prompt}
\end{figure*}

\begin{figure*}[t]
\begin{tcolorbox}[
    listing only,              
    colback=black!5,
    colframe=black!75,
    arc=2mm,
    boxrule=0.5pt,
    width=\textwidth,
    left=4mm, right=4mm, top=2mm, bottom=2mm,
    listing options={
        basicstyle=\ttfamily\small,
        breaklines=true,        
        showstringspaces=false, 
        columns=flexible,
        escapechar=|,
    },
]
You are an expert AI Safety Adjudicator. We are classifying user-model interactions to identify refusal behaviors. Analyze the [User Query] and [Model Response] and classify them into one of the following three categories.

\textbf{[IMPORTANT RULES]}

Check if the Model Refused: \\
Did the model explicitly decline to answer? (e.g., ``I cannot fulfill this request,'' ``I apologize,'' ``I am unable to,'' or citing safety guidelines).
\begin{itemize}
    \item IF NO (Model Answered) $\rightarrow$ Output \texttt{[[2]]}.
    
    \item IF YES (Model Refused) $\rightarrow$ Proceed to check the User Query Safety:
    
    \begin{itemize}
        \item Is the User Query HARMFUL / SENSITIVE? \\
        (e.g., Illegal acts, violence, self-harm, sexual content, PII, hate speech). \\
        $\rightarrow$ Output \texttt{[[1]]} \\
        \textit{(Meaning: The query was dangerous, so the refusal was correct.)}
        
        \item Is the User Query SAFE / BENIGN? \\
        (e.g., General knowledge, harmless requests, creative writing, or incorrectly flagged topics). \\
        $\rightarrow$ Output \texttt{[[0]]} \\
        \textit{(Meaning: The query was safe, but the model refused. This is a False Refusal.)}
    \end{itemize}
\end{itemize}

\textbf{[INPUT INFO]}

<user query>: \{user query\}

<model response>: \{model response\}

\textbf{[OUTPUT FORMAT]}

Please output ONLY [[0]], [[1], or [[2]] without any markdown or explanation.
\end{tcolorbox}
\caption{The prompt used for the Refusal Validation step in \cref{sec:data_construct}. We find that gpt-oss-120b with this prompt can effectively identify justified refusals.}
\label{fig:app_refual}
\end{figure*}

\section{Experimental Details}
\label{sec:app_exp_detail}

We conduct all experiments using NVIDIA H100 GPUs. For \textbf{reward model training}, we adopt the huggingface TRL framework\footnote{\url{https://github.com/huggingface/trl}}. We adopt a batch size of $512$, a learning rate of $1 \times 10^{-5}$, and a maximum sequence length of $4,096$, training for a single epoch. Regarding reward model evaluation, we employ the RewardBench codebase\footnote{\url{https://github.com/allenai/reward-bench}} to assess performance on RewardBench, RM-bench, and JudgeBench and use the official evaluation code for PPE\footnote{\url{https://github.com/lmarena/PPE}}.
Regarding the \textbf{cross-sample consistency} pointwise evaluation in \cref{sec:consistency}, we first sample $5,000$ instances from the WildChat held-out set. Following the pipeline described in \cref{sec:data_construct}, we instead simplify the task from four-category (excluding the \textit{Neutral Ambiguity} category) to binary classification, i.e., positive and negative, to reflect the real user satisfaction (accept or reject) with the response.
We then downsample the negative instances for label balancing and collect about $1,000$ samples for manual verification. This verification process evaluates negative feedback to distinguish between objectively low-quality responses and those reflecting personal preference (which will be marked as noise). Two authors independently check the data, and we filter out any instance flagged as noise by at least one author. This results in a final set of $948$ instances.
Regarding \textbf{DPO training}, we construct a dataset of $20,000$ samples from Infinity-Instruct~\citep{li2025infinity}. Specifically, we adopt gpt-oss-120b~\citep{agarwal2025gpt} to rate instruction difficulty on a scale of 1 to 5, discarding samples with extreme scores (1 and 5). The final dataset contains a mix of $60\%$ subjective tasks, $20\%$ mathematics, and $20\%$ commonsense reasoning. We conduct training using the verl framework\footnote{\url{https://github.com/volcengine/verl}} and we integrate a custom Online DPO feature. We employ a batch size of $64$ with $8$ rollouts per prompt, a learning rate of $5 \times 10^{-7}$, and a maximum sequence length of $4,096$. For policy evaluation, we adopt \texttt{gpt-4o-mini-2024-07-18} as the judge model for Alpaca Eval 2.0 and Arena Hard. For IFEval, we report the average accuracy across the prompt strict, prompt loose, instruction strict, and instruction loose metrics.


\end{document}